# Computer Vision and Abnormal Patient Gait Assessment a Comparison of Machine Learning Models

Jasmin Hundal MD University of Connecticut, Benson A. Babu MD MBA Saint Johns' Episcopal Hospital, Plainview Medical Center Northwell Health

## Abstract

Abnormal gait, its associated falls and complications have high patient morbidity, mortality. Computer vision detects, predicts patient gait abnormalities, assesses fall risk and serves as clinical decision support tool for physicians. This paper performs a systematic review of how computer vision, machine learning models perform an abnormal patient's gait assessment. Computer vision is beneficial in gait analysis, it helps capture the patient posture. Several literature suggests the use of different machine learning algorithms such as SVM, ANN, K-Star, Random Forest, KNN, among others to perform the classification on the features extracted to study patient gait abnormalities.

## 1.0. Introduction

Gait abnormalities falls and associated complications have high morbidity and mortality. Monitoring of patient mobility may identify fall risk and potential treatment options. Advanced computer vision algorithms and more efficient low cost sensors prevent patient falls and its complications. Such preventable complications include pulmonary emboli, myocardial infarction, hip fractures and medical deconditioning. Thus, patient gait assessment matters in medicine, and several research works performed to analyze patient gait patterns. [1]. Gait associated with a stroke, dementia, frail, elderly examined in neurology, physical medicine rehabilitation, rheumatology, and orthopaedics regularly [2]. Analysis of gait, its parameters' performed routinely in clinical practice [3].

Gait, which is a part of humans' common behaviour, could show mental disorders like depression [4] dementia [5], intellectual disability [6], and musculoskeletal disorders like deformation of the joint [7]. The most significant factor used in carrying out and assessing the patient's therapy and rehabilitation is the accurate measurement of the amount of exercise performed during everyday life. As a result, there are several bodies of research on gait [8].

The systematic study of human walk, its abnormalities, theories for its etiology, recommended treatment, is gait analysis. Clinical applications usually use gait analysis in identifying medical conditions or for monitoring the state of the patients' clinical recovery [9]. Clinicians observe gait patterns of a patient as he or she walks are one's responsible for carrying out the medical treatment plan. Gait analysis is a subjective, determined by the clinicians' experience and judgment. Because of the subjective nature of gait analysis, it impacts diagnosis and treatment decisions, thus patient outcomes [10].

Therefore, this project performs a systematic review of how computer vision used along with machine learning models to perform effective abnormal patient gait assessment. The research questions identified for this research are:

- **RQ 1:** What are the machine learning models used for the gait analysis?
- **RQ 2:** How can computer vision assist in gait analysis?

We organise this paper in multiple sections: Section 2 discuss the background study performed. Section 3 discuss the method adopted to perform the literature review. Section 4 shows the findings and analysis from the systematic review performed. Finally, section 5 discusses the conclusion of the work, along with suggested future works.

## 2.0. Background Study

This section provides a background study on the gait analysis, computer vision and machine learning concepts.

## 2.1. Gait Analysis

The definition for gait [11] :

*"The anthropomorphic upstanding self-displacement, in the stepping of two feet which alternates and that is with no extra fulcra, and that always contains support on a slightly inclined or horizontal surface."*

The scientific study of animal locomotion, especially those of humans referred to as Gait Analysis [12], and it primarily aims at determining the functions, categorisation and inconsistencies associated with gait, to provide better treatment for ambulatory patients. Gait analysis uses several approaches, including medical imaging technique, acoustic tracking system, magnetic system, goniometric measurement system, electromyography, foot planter presser sensor, force shoes, force plate mechanism, inertial system, optical system and utilities portable devices.

## 2.2. Computer Vision

Human vision refers to gazing at the world to understand it. Computer vision is similar; it, however, differs, because it uses a machine, especially a camera, for getting information. We use the following features for classifying a computer vision's entire applications [13]:

- **Gauging:** It relates to tolerance checking and dimensional characteristic measurement
- **Sorting:** It relates to the recognition and identification of parts.
- **Inspection:** It relates to detecting, identifying and classification of parts.

Within the past decade, they have conducted extensive research on video-based human motion capture. Various techniques in machine learning and computer vision are proposed for pose estimation and 3D human motion tracking [14]. A video-based technique used for carrying out joint kinematics assess while gait is ongoing developed by the work of Corazza et al. [15].

## 2.3. Machine Learning

As deep learning approaches emerge and advance, DNN-based techniques are the standard in visions tasks such as human motion tracking and pose estimation [16], human activity recognition [17] and face recognition [18]. Several hidden layers between the output and input layers, and the ones that can learn semantic and high-level features from the data to model complex non-linear relationships, make up DNNs. The current techniques on DNN for 3D human pose estimation focus on a single view, and a complex setting [14] [19]. Regarding clinical gait analysis, techniques in machine learning, which includes Logistic Regression [20], Artificial Neural Networks (ANN) [21], K-Star [22], Random Forest [23], K-nearest neighbours (KNN) [24] and Support Vector Machines (SVM) [25] found in applications identify and classify specific gait patterns into the medical conditions [16].

## 2.4. AI models for Gait Analysis

There are several Artificial Intelligence (AI) models introduced in the literature to identify various abnormalities in a particular patient by pose estimation and gender recognition. The following summarizes the AI models used for handling abnormalities:

### 2.4.1. Post Estimation

The human body's trunk and joints are parts, detected with human pose estimation [26]. The human pose estimation can detect these parts using videos or images from a video or image detector. Key points used to describe details of the human skeleton using human pose estimation [27]. For instance, the key points of a human skeleton's coordinates generated using a pose estimation model, uses the human body's photos as inputs. The role of the human pose estimation is pivotal in the prediction of human behaviour and a description of the human posture. Human pose estimation is among computer vision's basic algorithm and is pivotal in several related domains, including gait recognition, character tracking, action recognition and behaviour recognition [28].

The pictorial structures model [29], [30], [31] that expresses spatial relationships among the body parts as a kinematic-priors-based tree-structured graphical model, which couples connected limbs, made up the classical approach to articulated pose estimation. These methods can make typical mistakes, including counting image evidence twice, which can take place because of the connections between variables that a tree-structured model did not capture, even though they have been successful on images, in which the person's entire limbs. They used the pictorial structure model in the work of [32]; however, its underlying graph representation is different.

Hierarchical models [33] [34] signify how the parts relate at various sizes and scales in a hierarchical tree structure. Usually, the image structure easily detected, facilitates the location of harder-to-detect and smaller parts can be discriminative in the larger parts corresponding to the limbs, based on these models' underlying assumption.

Interactions which introduce loops so it can augment the tree structure with more edges capturing long-range, occlusion, and symmetry correlations, incorporated in non-tree models [35] [36]. Typically, the approximate inference required in these methods for learning, and test time. As a result, spatial relationships' accurate modelling traded off with models which enable efficient inference, mostly with a straightforward parametric form, to ensure fast inference.

Contrariwise, sequential-prediction-framework-based approaches [37] use likely complicated correlations between variables, for learning an implicit spatial model; they train an inference procedure directly, to achieve that, as [38] [39] shows.

The models that carry out articulated pose estimation [40] [41], using convolutional architectures have received much attention in recent times. The method that [42] used involves the use of of a standard convolutional architecture to carry out a direct regression of the Cartesian coordinates [43]. It regresses image to confidence map in recent work and opts for graphical models that must use spatial probability priors' heuristic initialisation or energy functions designed by hand, for removing outliers on the regressed confidence maps. It uses a dedicated network model in some of them for precision refinement [44], [45]. This paper shows that it is suitable to input the regressed confidence maps to additionally convolutional networks that do not require using hand-designed priors, that has great receptive domains for learning and attains high-level performance within the entire precision region. Further, it should not be carefully initialised and should have dedicated precision refinement. A network module with a large receptive field is used in the work of [46] for capturing implicit spatial models. [41] considered joint training's advantages; hence, the model we proposed can be trained globally because of convolutions differentiable attributes.

A deep network with the features of being able to use error feedback for training is seen in the work of [47]. It also uses Cartesian representation, as seen in [48] that is incapable of preserving spatial improbability, and that reduces the high precision regime's accuracy.

Carrying out the task of articulated pose estimation using Convolutional Pose Machines (CPMs). Convolutional Pose Machines (CPM) inherit pose machine architecture's benefits [37];integrating learning and inference tightly, the learning of long-range dependencies between multi-part cues and image implicitly, and a modular sequential design. It combines these with the benefits which convolutional architecture provides. CPMs also include advantages such as the capability of handling large training datasets efficiently, a differentiable architecture which makes joint training with backpropagation possible, and the ability to learn spatial and image context's feature representations directly from data. Series of convolutional networks, which constantly 2D maps for each part's location, make up CPMs.

CPMs [49] are robust, and they have a very high accuracy of detection on human pose estimations' standard datasets, such as Leeds Sports Pose (LSP) data set [50], Human Pose data set [51], and Max-Planck-Institut Informatik (MPII). The time required for training in CPMs is extensive, and its detection speed is low. This makes it challenging to be applied in real-time tasks. Based on human pose estimation's standard datasets, excellent detection outputs are found in the Stacked Hourglass [52] of similar duration. The new modes using the enhanced Stacked Hourglass include the 2017 models such as Learning feature [53], Self-Adversarial Training [54] and Multi-context [55], and the 2018 excellent models' further improved accuracy. Some metrics and contained in these models, and they elongate the time required for training, thereby making up a common limitation. To date, the model does not have a satisfactory accuracy.

The ability to extract the low-level feature is enhanced, using the more convoluted network structures, and deeper network layers of the enhanced CPM model [56]; and afterwards, apply a system to fine-tune it. The enhanced CPM is proven to include an excellent image detection effect and high image classification accuracy, and a good human pose estimation model for designing a new network and apply a system of fine-tuning to increase the human pose estimation's efficiency.

### 2.4.2. Gender Recognition

Using Gait Energy Image (GEI), which is a combination of gait and a new spatiotemporal method for force representation, for marking human walking's behaviour for individual recognition, proposed in the work of Han et al. The findings of the study shows the efficiency of the use of the combination of gait and GEI approach for individual recognition, and the competitiveness of its performance [57]. [58] used the GEI approach for studying individual recognition. They used various techniques and outlooks to present the GEI approach as biased attributed in their survey. It is clear from the findings of their research that the system's performance in real-time improved; hence, its application in real-word is possible. [59] used automated approaches to combine psychological methods for improving accuracy quality, to classify human gait-based genders. According to the research, compared to other parts of the body, the major body parts for gender recognition process include the chest, back, hair and head. Even though the application process contains several impediments are because of the differences in how humans appear, and they include change of shoes and clothes, or when they lift objects, the gait classification is possible in a controlled environment.

The classification of human behaviour using 2-Directional 2-dimensional principles component analysis ((2D)2PCA) and 2G (2D) 2PCA) Enhanced Gait Energy Image (EGEI) proposed in the work of [60]. The outcomes of the experiment revealed the simplicity of the algorithm and its capacity for realising a higher classification accuracy within a short period. A system that can use gait classification based on the silhouette, to recommend books to visitors according to their age or gender, and in real-time proposed in the work of [61].The Super Vector Machine (SVM) approach with 77.5% accuracy rate, used in the classification process. [62] combined the denoised energy image (DEI) and GEI approach in the pre-processing process to present gender recognition's initial design and outcomes from experiment from walking movements. The training of the process, and the extraction of the feature, used the Support Vector Machine (SVM). It is clear from the research's findings that certain typical values will used in the proposed approach; it could be 100 percent accurate.

The method of integrating information from the multi-view gait at the feature level proposed in the work of [63], and it increases the effectiveness of the performance for the gender classification based on multi-view gait. A study on the use of gait for human recognition was conducted in the work of [64]. Gait image's features that are founded on information theory sets referred to as image feature information gait are presented in this report. Gait information features, which are information set theory-based gait image features are presented in this report. The concept of the information set was applied on the frames in a gait cycle, and two elements referred to as Gait information image with Sigmoid feature (GIISF) extracted and Gait information Image with Energy Feature (GII-EF) to derive the proposed Gait information Image (GII). The identification of the gait was made using Nearest Neighbour (NN) for the classification. The robust feature-level fusion of directional vectors such as forward and backward diagonal, vertical, and horizontal vectors used in the work of [65] to study gender recognition. The first construct for each image sequence was Gait Energy Image (GEI), followed by Gradient Gait Energy Image (GGEI), which is achieved using neighbourhood gradient computation. After that, differences in all the four directions were utilised as discriminative gait features. Afterwards, SVM used in the classification process, while the largest multi-view CASIA-B (Chinese Academy of Sciences) datasets were significantly used in the testing process. The investigators noted that the outcome could be potentially beneficial.

According to the literature review, the current most universal gait-based approaches to gender classification include GEI and GII approaches. As a result, this research focuses on contrasting GII approaches with GEI approaches to present a gait-based gender classification in real-time. The one with the highest accuracy will be beneficial for future study.

### 3.0. Methods

### 3.1. Search criteria

The systematic review aimed at reviewing published papers, as well as academic journals in a step yby step manner. It also intends to perform a systematic peer-review on academic-based journals. It will use online search engines such as IEEExplore[1], PubMed[2], Google Scholar[3], Cochrane[4], CINAHL, Medline[5], Web of science[6], DBLP[7], and EMBase[8] to search for literature. The primary keywords used for the search are Computer vision, Artificial Intelligence, Machine learning, Deep learning, CNN, Abnormal gait analysis, gait analysis, Stroke, Parkinson's Disease, and Movement disorders.

### 3.2. Justification of the selection

The preliminary research produced one hundred articles. We considered only ten of them in this report. Out of the ten articles, only five of them related to this report's topic. The report set duration of from between 2009 and 2019 for the works of literature, to ensure that only up-to-date works of literature used. However, sometimes, some earlier journals used.

## 4.0. Findings and Analysis

The key findings from the journals are provided in table 1.

*Table 1: Key findings*

| System suggested | Year | Computer Vision technology used | Machine learning technique used | The abnormality identified using the Gait analysis | Reference |
|---|---|---|---|---|---|
| Automatic Health Problem Detection | 2018 | Videos captured using digital cameras | DNN | Parkinson's disease Pose Stroke orthopaedic problems | [16] |
| A vision-based proposal for classification of normal and abnormal gait | 2016 | RGB Camera | KNN and SVM | Dementia frailty | [2] |
| Computer Vision-Based Gait Analysis | 2018 | Smart Phone | KNN | Senility Frailty | [2] |
| Extracting Body Landmarks from Videos | 2019 | Videos | Suggested future work for classification or regression algorithms | Parkinson disease | [66] |
| System to support the Discrimination of Neuro-degenerative Diseases | 2009 | Videos | SVM, Random Forest and KStar | Amyotrophic lateral sclerosis, Parkinson's disease and Huntington's disease | [67] |

Several measures identified in the gait analysis to study the abnormality of the patients, some of which we provide in table 2.

*Table 2: The measures identified in the gait analysis to study the abnormality of the patients [68]*

| Patient Abnormality | Gait measures |
|---|---|
| Slow walking | Gait speed Frequency of steps |
| Muscle weakness | Muscle force |
| Crouch Gait | Ankle joint angle |
| Unstable gait | Gait stability measure Double support time |
| High stepped gait | Step height |
| Pelvis drop | Hip flexion |

## 5.0. Conclusion

From this study, it is shows several machine learning algorithms suggested in the literature perform the classification, include SVM, K-Star, Random Forest, KNN and DNN. The images and videos widely used in the literature to capture the human walk while performing the gait analysis. Therefore, the use of high technologies of computer vision, such as smartphone cameras, surveillance cameras, among others, are increasing drastically. This research's limitation includes its failure to perform in-depth research on the gait analysis and its functions. The comparison of the approach used to perform the gait analysis in the literature performed to know the in-depth information regarding the gait analysis and its usage.

---

[1] https://ieeexplore.ieee.org/Xplore/home.jsp
[2] https://www.ncbi.nlm.nih.gov/pubmed/
[3] https://scholar.google.com/
[4] https://www.cochranelibrary.com/
[5] https://www.ebsco.com/products/research-databases/medline
[6] https://clarivate.com/webofsciencegroup/solutions/web-of-science/
[7] https://dblp.uni-trier.de/
[8] https://www.embase.com/login